\documentclass[cameraready]{Interspeech}
\usepackage{booktabs}
\usepackage{multirow}
\usepackage{pifont}
\usepackage{makecell}
\usepackage[T1]{fontenc}

\title{Cross-modal Consistency Guidance for Robust Emotion \\ Control in Auto-Regressive TTS Models}
\author[affiliation={1,2}]{Yizhou}{Peng}
\author[affiliation={3}]{Yukun}{Ma}
\author[affiliation={3}]{Chong}{Zhang}
\author[affiliation={2}]{Yi-Wen}{Chao}
\author[affiliation={3}]{Chongjia}{Ni}
\author[affiliation={3}]{Bin}{Ma}
\author[affiliation={2}]{Eng Siong}{Chng}

\address{
    $^1$ Alibaba-NTU Global e-Sustainability CorpLab, Nanyang Technological University, Singapore \\
    $^2$ College of Computing and Data Science, Nanyang Technological University, Singapore \\
    $^3$ Alibaba, Alibaba Inc., Singapore
}

\email{peng.yizhou@ntu.edu.sg}

\keywords{Emotional Speech Synthesis, Classifier-Free Guidance, Reinforcement Learning}

\usepackage{comment}

\begin{document}

\maketitle

\begin{abstract}
    While Text-to-Speech (TTS) systems enable emotional control via natural-language instructions, expressiveness, naturalness, and speech quality degrade when the target emotion conflicts with the textual semantics. We propose a Cross-modal Consistency Guided Classifier-Free Guidance (CCG-CFG) method with dynamic scales based on the degree of inconsistency between the text emotion and the explicit speech emotion, replacing the dropout condition with the text emotion. We also distill the CCG-CFG guidance signal using a hard-sample mining strategy, improving the TTS model's emotional alignment capability.
    Evaluations on five emotional corpora and two TTS benchmarks show that our approaches applied to CosyVoice2 achieve up to a 12\% absolute improvement in emotion-recognition accuracy and a 10\% relative improvement in subjective scores, outperforming baselines including HierSpeech++, Qwen3-TTS, and the original CosyVoice2, while preserving intelligibility, naturalness, and high speech quality.
\end{abstract}

\section{Introduction}

Modern Text-to-Speech (TTS) systems are increasingly expected to produce not only intelligible but also highly expressive and emotionally resonant speech for applications such as virtual assistants, audiobook narration, and digital avatars~\cite{xie2024towards,clifton2020100}. With the advent of Large Language Models (LLMs), the paradigm of emotional TTS has fundamentally shifted~\cite{yang2023uniaudio,ma2025review}. By leveraging deep semantic understanding, modern systems can parse complex prompts~\cite{he2024can,du2024cosyvoice,huang2025step}, transitioning from coarse-grained predefined labels (e.g., ``happy'') or reference audio snippets~\cite{jawaid2024style,cho2024emosphere} to nuanced natural language instructions (e.g., ``speak in a calm and reassuring tone''). This shift enables unprecedented zero-shot and fine-grained controllability over the synthesized speech~\cite{lee2025hierspeech++,jing2025enhancing}.

However, a critical challenge arises in realistic scenarios: \textbf{A cross-modal inconsistency between the text and rendered emotion in synthesized speech.} As users are empowered to provide arbitrary textual semantics alongside explicit emotion instructions, there is frequently an inconsistency, or even direct conflict between the \textbf{text emotion} and the \textbf{rendered emotion} (e.g., requiring a TTS to say ``It isn't a happy memory'' in a \textit{surprised} tone). When faced with these cases, the model must reconcile competing signals while maintaining naturalness, intelligibility, and expressive fidelity. 
Consequently, robust emotional TTS requires not only accurate speech modeling but also dedicated mechanisms to resolve this cross-modal inconsistency. This challenge remains largely underexplored in auto-regressive (AR) TTS frameworks~\cite{hussain2025koel,Qwen3-TTS,du2024cosyvoice}, the current state-of-the-art for natural, zero-shot speech synthesis. To enhance style alignment in these models, a common workaround is to apply Classifier-Free Guidance (CFG)~\cite{ho2021classifier,liu2023audioldm,zheng2023guided} during inference. By extrapolating the conditional prediction away from an unconditional baseline, CFG artificially amplifies the target style; however, it naively introduces severe synthesis artifacts~\cite{parakeet-tts}.


This work\footnote{Demo Page: \scriptsize\url{pengyizhou.github.io/Emotional_tts_demo}} addresses the cross-modal emotion inconsistency challenge with a Cross-modal Consistency Guided CFG (CCG-CFG)  
scheme that modulates this emotional non-congruence.
Specifically, our contributions lie in the following:
\begin{itemize}
 \item We propose a CCG-CFG framework that replaces the CFG's \textbf{unconditional dropout} with the \textbf{text emotion} if inconsistency is detected, which explicitly magnifies the distance between the conditional and unconditional predictions to provide a precise and more effective guidance signal.
 \item We further improve the CCG-CFG with a \textbf{dynamic guidance scale} (DS-CCG-CFG) based on the LLM-measured degree of inconsistency, 
 which balances strong emotional expressiveness with speech naturalness and intelligibility.
\item Finally, we distill the DS-CCG-CFG signal into the AR TTS model via an \textbf{inconsistent-sample mining strategy}, removing the two-pass decoding overhead and CFG artifacts.


\end{itemize}

\section{Methods}
\label{sec:methods}
\subsection{Overall Architecture}
\begin{figure}[th]
    \centering
    \includegraphics[width=0.9\linewidth]{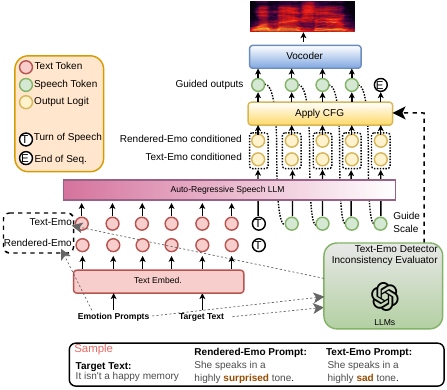}
    \vspace{-1.0em}
    \caption{Overall architecture of the proposed Dynamic Scale Cross-Modal Consistency Guided CFG (DS-CCG-CFG) framework. The guidance mechanism replaces standard unconditional dropout with the  \textbf{Text-Emo} to contrast against the \textbf{Rendered-Emo} condition. An LLM evaluator determines the text emotion and the degree of cross-modal inconsistency, dynamically mapping this level to an appropriate guidance scale for the AR TTS model.}
    \label{fig:architecture}
    \vspace{-2.0em}
\end{figure}

Figure~\ref{fig:architecture} illustrates the inference pipeline of our proposed CCG-CFG framework integrated into an AR TTS model. Given a target text and a user-specified \textbf{Rendered-Emo} prompt, an external LLM first extracts the \textbf{Text-Emo}. Simultaneously, the LLM assesses the degree of cross-modal inconsistency between these two emotional states, produces an \textbf{inconsistency profile} in \{\textit{Identical, Inconsistent, Highly Inconsistent}\}, and dynamically maps this mismatch onto a specific guidance scale. 

During the inference phase, the model runs two parallel passes: one conditional on Rendered-Emo and the other on Text-Emo instead of the dropout condition of the traditional CFG, whose logits are fused by the dynamic scale $w$. By contrasting the two emotions, CCG-CFG yields an inconsistency-aware signal that steers token generation before vocoding.

\subsection{Classifier-Free Guidance}
We first demonstrate that cross-modal inconsistency fundamentally impairs AR TTS generation by assessing synthesis quality under varying degrees of text-rendered emotion inconsistency. 
As shown in Table~\ref{tab:mismatch}, TTS baselines (CosyVoice2 and HierSpeech++) suffer severe EmoACC degradation and a slight increase in Word Error Rate (WER) when rendered emotions conflict with textual semantics. Ground-truth recordings, in contrast, carry the rendered emotion as a genuine acoustic property; since the SER model classifies acoustic rather than textual affect, its EmoACC remains high regardless of text-rendered inconsistency.
\begin{table}[htp]
\centering
\footnotesize
\caption{Emotion Expressiveness and intelligibility results by Model and \textbf{Incon}sistency profile. \ding{55} means that the text and rendered emotions are identical, while \ding{51} and \ding{51}\ding{51} stand for inconsistent and highly inconsistent between the two emotions. HierSpeech++ and CosyVoice2 use Ground Truth audio as reference speech for synthesizing.}
\renewcommand{\arraystretch}{0.8}
\begin{tabular}{lccc}
\toprule
\textbf{Models} & \textbf{Incon.} & \textbf{EmoACC(\%)$\uparrow$} & \textbf{WER(\%)$\downarrow$} \\
\midrule
\multirow{3}{*}{Ground Truth}
  & \ding{55}        & 91.16 & 5.62 \\
  & \ding{51}        & 85.88 & 5.91 \\
  & \ding{51}\ding{51} & 82.33 & 6.80 \\
\midrule
\multirow{3}{*}{HierSpeech++}
  & \ding{55}        & 80.32 & 4.52 \\
  & \ding{51}        & 58.44 & 4.53 \\
  & \ding{51}\ding{51} & 48.82 & 5.11 \\
\midrule
\multirow{3}{*}{CosyVoice2}
  & \ding{55}        & 80.92 & 4.76 \\
  & \ding{51}        & 71.77 & 4.43 \\
  & \ding{51}\ding{51} & 63.70 & 4.84 \\
\bottomrule
\end{tabular}
\label{tab:mismatch}
\vspace{-2.0em}
\end{table}

\subsubsection{Cross-modal Consistency Guided CFG (CCG-CFG)}


Standard CFG attempts to enforce alignment by emphasizing a conditional prediction (rendered-emo, $c_{re}$) from an unconditional prediction ($\emptyset$), stated as Eq.~\ref{eq:ca_cfg_consistent}, where $w>1.0$ is the guidance scale. However, merely dropping the condition is insufficient to overcome the strong semantic pull of the inconsistency.
To explicitly expand the gap between the two conditions, CCG-CFG replaces $\emptyset$ with $c_{te}$ as the \textit{text emotion} when \textbf{cross-modal inconsistency is detected}, calculated as Eq.~\ref{eq:ca_cfg_conflict}, while emotion identical cases stick to Eq.~\ref{eq:ca_cfg_consistent}.

\vspace{-1.0em}
\begin{align}
    L_{CFG} &= L(\emptyset) + w \cdot (L(c_{re}) - L(\emptyset)) \label{eq:ca_cfg_consistent} \\
    L_{CCG-CFG} &= L(c_{te}) + w \cdot (L(c_{re}) - L(c_{te})) \label{eq:ca_cfg_conflict}
\end{align}

\subsubsection{Dynamic-Scale CCG-CFG (DS-CCG-CFG)}
\begin{figure}[htp]
    \centering
    \vspace{-0.5em}
    \includegraphics[width=0.9\linewidth]{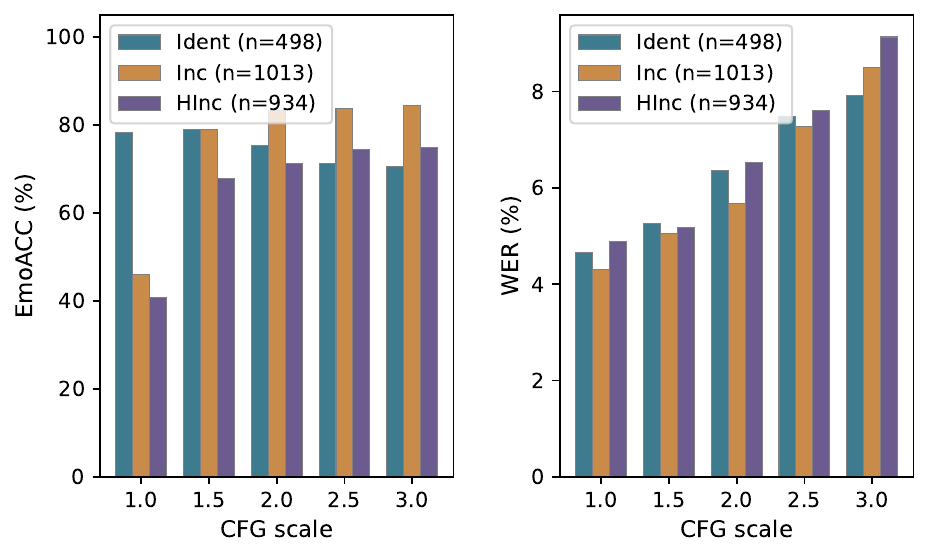}
    \vspace{-1.0em}
    \caption{EmoACC and WER by CFG scale for three inconsistency profiles: \textbf{Ident}ical; \textbf{Inc}onsistency, and \textbf{H}ighly-\textbf{Inc}onsistency. $Scale=1.0$ stands for no guidance applied.}
    \label{fig:cfg_scale}
\vspace{-1.5em}
\end{figure}

CFG usually applies a fixed guidance scale $w$, and this is sub-optimal~\cite{betaCFG,dynamicCFG}.
As illustrated in Figure~\ref{fig:cfg_scale}, CFG exhibits divergent behavior depending on inconsistency severity: guiding \textit{Identical} samples degrades EmoACC, while moderate guidance improves expressiveness for \textit{Inconsistent} samples, excessively large scales severely compromise intelligibility, leading to a sharp spike in WER across all conditions. 
To resolve this, DS-CCG-CFG dynamically adjusts $w$ based on the three detected inconsistency profiles and maps them to the scales of \{1.0, 2.5, 3.0\}, respectively. These scale values were selected via a grid search on the development set,
balancing EmoACC against WER for each profile.
\subsection{Guidance Distillation and Hard Sample Mining}
To eliminate the inference overhead of two-pass decoding and the synthesis artifacts introduced by CFG, we distill the DS-CCG-CFG guidance signal directly into the TTS model via Direct Preference Optimization (DPO)~\cite{DPO,emodpo,dpo-tts-llm}. Furthermore, we maximize distillation efficacy using a hard-sample mining strategy: for each text in an external corpus, we extract its text emotion and pair it with a contrasting target emotion to artificially induce a cross-modal inconsistency corpus. 

To construct the DPO preference pairs, we generate 25 candidate speech samples per target text by applying five random seeds across five CCG-CFG guidance scales ($w \in \{1.0, 1.5, 2.0, 2.5, 3.0\}$). These candidates are then ranked using an objective score $S$ that balances intelligibility and expressiveness:
\vspace{-0.5em}
\begin{equation}
S = 0.5 \cdot (1-\text{WER}) + 0.5 \cdot \text{EmoConf}
\end{equation}
where $\text{EmoConf}$ adjusts the raw emotion prediction confidence $p$ by applying a penalty if the predicted emotion $\hat{y}$ does not match the target emotion $y$:
\vspace{-0.5em}
\begin{equation}
\text{EmoConf} = 
\begin{cases} 
p, & \text{if } \hat{y} = y \\ 
p - 1, & \text{otherwise} 
\end{cases}
\end{equation}
\vspace{-1.0em}

By selecting the highest- and lowest-scoring candidates as positive and negative pairs, this approach compels the model to learn stable emotion control under cross-modal inconsistency while preserving performance for emotion-identical cases.
Because the objective score $S$ enforces an intelligibility-and-quality constraint when ranking candidates, CFG-induced artifacts are pushed toward the rejected side and filtered out during preference optimization, rather than being distilled into the model.

\begin{table}[htbp]
\centering
\footnotesize
\vspace{-0.5em}
\renewcommand{\arraystretch}{0.8}
\caption{Data distribution across the combined dataset and VCTK external resource (used as additional data for RL training). The EXT duration is not reported because we only use the text and perform speech synthesis for RL training.}
\begin{tabular}{lrrrrr}
\toprule
 & Train & Valid & Test & EXT. (text) \\
\midrule
\textbf{Total} & \textbf{37{,}883} & \textbf{2{,}445} & \textbf{2{,}445} & \textbf{20{,}000} \\
Duration (h) & 40.01 & 2.83 & 2.85 & -- \\
\# Speaker & 985 & 252 & 244 & 7 \\ 
\bottomrule
\end{tabular}
\label{tab:emotion-summary}
\vspace{-1.5em}
\end{table}

\section{Experimental Setup}
\label{sec:exp-setup}

\subsection{Dataset}
\label{sec:data}
Our experiments are conducted using a combined corpus sourced from seven well-established speech datasets: ESD~\cite{ESD}, MESS~\cite{MESS}, MEAD~\cite{MEAD}, TESS~\cite{TESS}, SAVEE~\cite{SAVEE}, LibriTTS~\cite{LIBRITTS}, and VCTK~\cite{VCTK}. We curated this dataset to cover seven distinct emotion categories: angry, disgusted, fearful, happy, neutral, sad, and surprised. 
Table~\ref{tab:emotion-summary} summarizes the data splits. 
The training set comprises 38k utterances (40 hours), while the validation and test sets contain 2.5k utterances (roughly 2.8 hours) each. We also have a supplementary 20k text-only samples from the VCTK dataset for the hard-sample mining strategy.

\subsection{Model Configurations}
\subsubsection{Baseline models}
To evaluate the effectiveness of our proposed framework, we compare it against several state-of-the-art emotional TTS systems:

\begin{itemize}
    \item \textbf{HierSpeech++}~\cite{lee2025hierspeech++}: A zero-shot speech synthesis framework. For emotion control, an explicit reference speech sample containing the rendered emotion is required.
    \item \textbf{Qwen3-TTS}~\cite{Qwen3-TTS}: A large-scale, LLM-based TTS system. While it supports direct control of emotion tags, this feature is limited to its built-in, optimized voices. Otherwise, it also requires a reference speech containing the rendered emotion.
    \item \textbf{CosyVoice2}~\cite{du2024cosyvoice}: An auto-regressive TTS system that natively supports both reference-based emotional transfer and Natural Language Emotion Control (NLEC). 
\end{itemize}

\subsubsection{Supervised Finetuning (SFT) and DPO training}
We fine-tune the CosyVoice2 LLM on the \textbf{Train} subset for SFT, and \textbf{Train} with \textbf{EXT} hard samples for DPO training. Each training is performed on a single H20-96GB GPU. The total effective batch size is 640s, with a constant learning rate of $10^{-5}$ and optimized using the Adam optimizer. The models are trained for 4 epochs without applying model averaging.

\subsubsection{Inference configurations on CosyVoice2}
We integrate the traditional CFG and the proposed CCG-CFG methods into the \textbf{CosyVoice2} TTS model, with two guidance scale $w$ configurations: $w=1.5$ and $w=3.0$, corresponding to \textbf{L}ow and \textbf{H}igh, respectively. 
We use \textit{gpt-5.2-2025-12-11} as the LLM to judge the text emotion and to discriminate cross-modality inconsistency. Note that the LLM provides only a
control signal for guidance scaling; it is not used for evaluation or as scientific
evidence, so our reported gains rest entirely on the independent objective and
subjective metrics.

\subsection{Evaluation Methods}

\subsubsection{Emotional Expressiveness}
We use a pre-trained Speech Emotion Recognition model (\textit{emo2vec-plus-large}~\cite{emo2vec}) and report the average \textit{EmoACC} to quantify how accurately synthesized speech conveys the target emotion.

\subsubsection{Speech Quality, Naturalness, and Intelligibility}
We evaluate the quality and naturalness of the synthesized audio using three objective metrics: \textit{UTMOS}~\cite{utmos} to predict the Mean Opinion Score (MOS) for speech naturalness, \textit{DNSMOS}~\cite{dnsmos} for audio clarity, and \textit{NISQA}~\cite{nisqa} for synthesized speech naturalness. 
To assess the clarity and correctness of the synthesized speech, we transcribe the generated audio using an ASR model (\textit{Whisper-Large-V3-Turbo}~\cite{whisper}) and compute the \textit{WER}. 


\begin{table*}[t]
\centering
\caption{Objective evaluation results across various model configurations. \textbf{NLEC} indicates support for Natural Language Emotion Control. \textbf{Rendered} \textbf{Emo}tion \textbf{Ref}erence denotes whether we provide the reference speech with the specific \textbf{R}endered or a fixed \textbf{N}eutral emotion. \textbf{(L)}ow and \textbf{(H)}igh denote low- and high-guidance scales in \textbf{CFG} and \textbf{CCG-CFG}, respectively, whereas \textbf{DS-CCG-CFG} applies the \textbf{Dynamic Scale} over CCG-CFG. \textbf{DS-CCG-CFG for DPO} is our proposed Guidance Distillation method, while \textbf{HARD EXT. DATA} is from the Hard Sample Mining method. \textbf{MaJ} mimics subjective evaluation using Model-as-Judge.}
\footnotesize
\renewcommand{\arraystretch}{0.8}
\begin{tabular}{l|cccccccc}
\toprule
\multirow{2}{*}{\textbf{Model \& Config}} & \multirow{2}{*}{\textbf{NLEC}}& \multirow{2}{*}{\textbf{\makecell{Rendered\\ Emo Ref}}} & \multicolumn{6}{c}{\textbf{Metrics}} \\
 \cmidrule(lr){4-9}
& & & \textbf{EmoACC (\%)$\uparrow$} & \textbf{WER (\%)$\downarrow$} & \textbf{UTMOS$\uparrow$} & \textbf{DNSMOS$\uparrow$} & \textbf{NISQA$\uparrow$} & \textbf{MaJ$\uparrow$} \\
\midrule

\textit{Ground Truth} & - & - & 85.60 & 6.04 & 2.27 & 2.91 & 3.48 & 65.4 \\
\midrule
HierSpeech++~\cite{lee2025hierspeech++}
  & \ding{55} & R\textsuperscript{$\dagger$} & 52.97 & 9.82 & 3.75 & 2.93 & 3.41 & 48.1 \\
\midrule

\multirow{2}{*}{Qwen3-TTS~\cite{Qwen3-TTS}} 
  & \ding{55} & R\textsuperscript{$\dagger$}  & 58.08 & 4.97 & 3.93 & 2.98 & 3.85 & 62.4 \\
  & \ding{51}\textsuperscript{$\ddagger$}  & -  & 49.98 & 4.21 & 3.83 & 3.09 & 4.36 & 80.9 \\
\midrule

\multirow{2}{*}{CosyVoice2} 
  & \ding{51} &  R & 65.11 & 4.74 & 3.72 & 3.06 & 3.44 & 55.8 \\
  & \ding{51} & N  & 50.63 & 4.61 & \textbf{4.32} & \underline{\textbf{3.24}} & \underline{\textbf{3.54}} & 53.0 \\
\midrule
  CosyVoice2-SFT & \ding{51} & N &
    53.54 &
    5.53 &
    4.23 &
    3.21 &
    3.57 &
    55.5 \\

\midrule
  CosyVoice2-CFG (L) & \ding{51} & N &
    52.43 &
    5.93 &
    4.26 &
    3.24 &
    3.54 &
    52.9 \\

  CosyVoice2-CFG (H) & \ding{51} & N &
    56.40 &
    9.88 &
    4.15 &
    3.22 &
    3.49 &
    50.8 \\
\midrule

\multicolumn{9}{l}{\textbf{\textit{Proposed Methods}} \textit{(Experiments conducted on CosyVoice2)}} \\ 

  CCG-CFG (L) & \ding{51} & N &
    55.50 &
    5.17 &
    4.27 &
    3.24 &
    3.53 &
    54.5 \\

  CCG-CFG (H) & \ding{51} & N &
    64.79 &
    9.17 &
    4.03 &
    3.21 &
    3.52 &
    57.8 \\
\midrule

DS-CCG-CFG 
  & \ding{51} & N & 
    \textbf{64.83} &
    7.86 &
    4.09 &
    3.22 &
    3.53 &
    \textbf{58.4} \\
\midrule

DS-CCG-CFG for DPO & \ding{51} & N &
    56.69 &
    \underline{\textbf{3.81}} &
    \underline{\textbf{4.31}} &
    \textbf{3.25} &
    \textbf{3.57} &
    56.3 \\

~~ +HARD EXT. DATA & \ding{51} & N &
    \underline{\textbf{59.55}} &
    \textbf{3.76} &
    4.29 &
    \textbf{3.25} &
    3.53 &
    \underline{\textbf{57.0}} \\


\bottomrule
\multicolumn{9}{l}{\textit{$\dagger$ \textbf{HierSpeech++} and \textbf{Qwen3-TTS} require a reference speech with the rendered emotion to perform emotional TTS without NLEC support.}} \\ 
\multicolumn{9}{l}{\textit{$\ddagger$ \textbf{Qwen3-TTS} supports emotion-tags with a built-in optimised voice. MaJ gives subscores close to 100 for naturalness and speech quality.}} \\

\end{tabular}
\label{tab:config_comparison}
\vspace{-1.5em}
\end{table*}
\subsubsection{Subjective Evaluation}
To complement our objective metrics, we conduct subjective evaluations using both human listeners and a Model-as-Judge (MaJ) approach. For the human evaluation, listeners are asked to rate the synthesized speech on a scale from 0 to 5 across three dimensions: Emotion Similarity to the ground truth (EMOS), Naturalness (NMOS), and Quality (MOS), using a 14-sample ($2 \times 7~\text{emos}$) subset. In parallel, we adopt a MaJ approach using the state-of-the-art multimodal \textit{Gemini-3.1-Pro-preview} to assess each synthesized sample on a scale of 0 to 100 across three dimensions: Emotional Expressiveness, Naturalness, and Intelligibility/Quality, across all experiment configurations. We report the mathematical average of these three scores as the final MaJ metric.

\section{Results and Analysis}
\label{tab:results}

\subsection{Evaluation Results}
As shown in Table~\ref{tab:results}, baselines such as \textit{HierSpeech++} and \textit{Qwen3-TTS} are optimized for emotion cloning, requiring rendered emotion references or relying on built-in voices with emotion tags. For NLEC-capable models such as CosyVoice2, performance degrades significantly when using a neutral rather than a rendered reference, where EmoACC drops from 65.11\% to 50.63\%. This suggests that providing an emotion reference leaks target-style information, thereby bypassing zero-shot NLEC. Consequently, all subsequent experiments use a neutral reference to mimic a realistic scenario, establishing CosyVoice2 without a rendered-emotion reference as our primary baseline. 
\begin{table}[tp]
\centering
\footnotesize
\caption{Emotion expressiveness and intelligibility results by Model and \textbf{Incon}sistency profile. \ding{55} means that the text and rendered emotions are identical, while \ding{51} and \ding{51}\ding{51} stand for inconsistent and highly inconsistent between the two emotions. CosyVoice2-N is CosyVoice2 with Neutral reference.}
\renewcommand{\arraystretch}{0.8}
\begin{tabular}{lccc}
\toprule
\textbf{Models} & \textbf{Incon.} & \textbf{EmoACC} & \textbf{WER} \\
\midrule
\multirow{3}{*}{CosyVoice2-N}
  & \ding{55}        & 78.31\% & 4.66\% \\
  & \ding{51}        & 46.00\% & 4.30\% \\
  & \ding{51}\ding{51} & 40.90\% & 4.89\% \\
\midrule
\multirow{3}{*}{DS-CCG-CFG}
  & \ding{55}        & 78.31\% & 4.66\% \\
  & \ding{51}        & 65.84\% & 7.91\% \\
  & \ding{51}\ding{51} & 56.53\% & 9.81\% \\
\midrule
\multirow{3}{*}{DS-CCG-CFG-DPO+EXT.}
  & \ding{55}        & 81.53\% & 3.19\% \\
  & \ding{51}        & 57.85\% & 3.78\% \\
  & \ding{51}\ding{51} & 49.68\% & 4.08\% \\

\bottomrule
\end{tabular}
\label{tab:mismatch_results}
\vspace{-3em}
\end{table}

Against this baseline, CosyVoice2-SFT partially recovers the lost expressiveness, improving zero-shot EmoACC (+2.91\%) with only a minor intelligibility penalty (WER +0.92\%).
CosyVoice2-CFG improves emotional rendering but degrades other metrics.
In contrast, CCG-CFG improves upon traditional CFG across all metrics by providing a more accurate cross-modal guiding signal.
The DS-CCG-CFG further optimizes this, delivering the best results among all training-free configurations by adaptively balancing expressiveness and overall quality, with the highest MaJ score of 58.4 and EmoACC of 64.83.

Finally, DS-CCG-CFG-DPO eliminates the reliance on external LLMs and two-pass CFG decoding during inference. This standalone DPO model outperforms zero-shot baselines and actively improves intelligibility, giving a lower WER of 3.81\%. Augmenting this training with an external corpus via hard-sample mining further boosts the EmoACC to 59.55\%, while preserving excellent naturalness and intelligibility, demonstrating robust emotion control without requiring inference-time guidance. 

\subsection{Cross-Modal Analysis and Human Evaluation}
Table~\ref{tab:mismatch_results} details performance under cross-modal inconsistency. The baseline CosyVoice2-N struggles with inconsistent samples; \textbf{DS-CCG-CFG} boosts EmoACC but introduces artifacts, while \textbf{DS-CCG-CFG-DPO+EXT.} resolves this trade-off. The largest gains occur on the incongruent samples, confirming that our distillation resolves cross-modal inconsistency without compromising quality.
\begin{table}[th]
  \caption{Human evaluation results. EMOS denotes Emotion Similarity to the ground truth, NMOS denotes Naturalness, and MOS denotes overall Speech Quality. -R and -N stand for using reference speech with Rendered emotion or Neutral emotion.}
  \label{tab:subjective_eval}
  \footnotesize
  \renewcommand{\arraystretch}{0.8}
  \centering
  \begin{tabular}{lccc}
    \toprule
    \textbf{Model } & \textbf{NMOS} & \textbf{EMOS} & \textbf{MOS} \\
    \midrule
    HierSpeech++       & 2.77 & 2.56 & 2.88 \\
    Qwen3-TTS-R          & \textbf{4.09} & \textbf{3.69} & \textbf{4.32} \\
    CosyVoice2-R         & 3.59 & 3.41 & 3.97 \\
    CosyVoice2-N  & 3.60 & 3.44 & 4.20 \\
    DS-CCG-CFG   & 3.31 & 3.43 & 3.91 \\
    DS-CCG-CFG-DPO+EXT. & \textbf{3.94} & \textbf{3.67} & \textbf{4.33} \\
    \bottomrule
  \end{tabular}
  \vspace{-1.0em}
\end{table}

Table~\ref{tab:subjective_eval} presents the human evaluation results. While applying \textbf{DS-CCG-CFG} degrades naturalness (3.31 NMOS) and quality (3.91 MOS) due to synthesis artifacts, \textbf{DS-CCG-CFG-DPO+EXT.} distilled from it successfully resolves this. By internalizing the guidance signal, the DPO model achieves the highest overall speech quality, with a MOS of 4.33. 
Furthermore, it yields substantial improvements in emotional expressiveness (3.67 EMOS) and naturalness (3.94 NMOS) over CosyVoice2-N and even outperforms the CosyVoice2-R, achieving performance comparable to the Qwen3-TTS-R.

\section{Conclusion}
This paper addresses the degradation of emotion TTS expressiveness and quality when the target emotion to be expressed in the synthetic speech conflicts with the textual semantics. To resolve this inconsistency, we propose a Cross-modal Consistency Guided Classifier-Free Guidance (CCG-CFG) scheme that dynamically scales guidance based on text-speech emotion inconsistency and conditions the unconditional pass on the text emotion. To eliminate the overhead of CFG decoding, we distill the guidance signal from the CCG-CFG directly into the TTS model, using a hard-sample mining strategy to enrich the training data.
Evaluations on CosyVoice2 demonstrate that our approach significantly improves emotion expressiveness, intelligibility, and naturalness, and it outperforms state-of-the-art baselines, including HierSpeech++ and Qwen3-TTS.

\section{Acknowledgement}
This research is supported by the RIE2025 Industry Alignment Fund - Industry Collaboration Projects (IAF-ICP) (Award I2301E0026), administered by A*STAR, as well as supported by Alibaba Group and NTU Singapore through Alibaba-NTU Global e-Sustainability CorpLab (ANGEL).
\section{Generative AI Use Disclosure}
The authors acknowledge the use of AI-assisted technologies to edit and polish the manuscript, improving language and clarity. All generated content was critically reviewed, and the authors assume full responsibility for the final manuscript.
\bibliographystyle{IEEEtran}
\bibliography{mybib}

@inproceedings{ESD,
  title={Seen and unseen emotional style transfer for voice conversion with a new emotional speech dataset},
  author={Zhou, Kun and Sisman, Berrak and Liu, Rui and Li, Haizhou},
  booktitle={ICASSP 2021-2021 IEEE International Conference on Acoustics, Speech and Signal Processing (ICASSP)},
  pages={920--924},
  year={2021},
  organization={IEEE}
}

@article{MESS,
  title={Categorical and dimensional ratings of emotional speech: Behavioral findings from the Morgan emotional speech set},
  author={Morgan, Shae D},
  journal={Journal of Speech, Language, and Hearing Research},
  volume={62},
  number={11},
  pages={4015--4029},
  year={2019},
  publisher={American Speech-Language-Hearing Association}
}

@inproceedings{MEAD,
  title={Mead: A large-scale audio-visual dataset for emotional talking-face generation},
  author={Wang, Kaisiyuan and Wu, Qianyi and Song, Linsen and Yang, Zhuoqian and Wu, Wayne and Qian, Chen and He, Ran and Qiao, Yu and Loy, Chen Change},
  booktitle={Computer Vision--ECCV 2020: 16th European Conference},
  pages={700--717},
  year={2020},
  organization={Springer}
}

@dataset{TESS,
  title={Toronto emotional speech set (TESS)},
  author={Dupuis, Kate and Pichora-Fuller, M Kathleen},
  year={2010},
  publisher={Scholars Portal Dataverse}
}

@misc{SAVEE,
  title={Surrey Audio-Visual Expressed Emotion (SAVEE) Database},
  author={Haq, Syed and Jackson, Philip JB and Edge, James},
  year={2008},
  howpublished={\url{http://kahlan.eps.surrey.ac.uk/savee/}},
  publisher={University of Surrey}
}

@inproceedings{LIBRITTS,
  title={LibriTTS: A Corpus Derived from LibriSpeech for Text-to-Speech},
  author={Zen, Heiga and Dang, Viet and Clark, Rob and Zhang, Yu and Weiss, Ron J and Jia, Ye and Chen, Zhifeng and Wu, Yonghui},
  booktitle={Interspeech},
  pages={1526--1530},
  year={2019}
}

@dataset{VCTK,
  title={CSTR VCTK corpus: English multi-speaker corpus for CSTR voice cloning toolkit},
  author={Yamagishi, Junichi and Veaux, Christophe and MacDonald, Kirsten},
  year={2019},
  publisher={University of Edinburgh. The Centre for Speech Technology Research (CSTR)}
}

@article{du2024cosyvoice,
  title={Cosyvoice 2: Scalable streaming speech synthesis with large language models},
  author={Du, Zhihao and Wang, Yuxuan and Chen, Qian and Shi, Xian and Lv, Xiang and Zhao, Tianyu and Gao, Zhifu and Yang, Yexin and Gao, Changfeng and Wang, Hui and others},
  journal={arXiv preprint arXiv:2412.10117},
  year={2024}
}

@article{huang2025step,
  title={Step-audio: Unified understanding and generation in intelligent speech interaction},
  author={Huang, Ailin and Wu, Boyong and Wang, Bruce and Yan, Chao and Hu, Chen and Feng, Chengli and Tian, Fei and Shen, Feiyu and Li, Jingbei and Chen, Mingrui and others},
  journal={arXiv preprint arXiv:2502.11946},
  year={2025}
}

@inproceedings{ho2021classifier,
  title={Classifier-Free Diffusion Guidance},
  author={Ho, Jonathan and Salimans, Tim},
  booktitle={NeurIPS 2021 Workshop on Deep Generative Models and Downstream Applications}
}

@article{jawaid2024style,
  title={Style mixture of experts for expressive text-to-speech synthesis},
  author={Jawaid, Ahad and Chandra, Shreeram Suresh and Lu, Junchen and Sisman, Berrak},
  journal={arXiv preprint arXiv:2406.03637},
  year={2024}
}

@inproceedings{xie2024towards,
  title={Towards controllable speech synthesis in the era of large language models: A systematic survey},
  author={Xie, Tianxin and Rong, Yan and Zhang, Pengfei and Wang, Wenwu and Liu, Li},
  booktitle={Proceedings of the 2025 Conference on Empirical Methods in Natural Language Processing},
  pages={764--791},
  year={2025}
}

@inproceedings{he2024can,
  title={Can large language models understand real-world complex instructions?},
  author={He, Qianyu and Zeng, Jie and Huang, Wenhao and Chen, Lina and Xiao, Jin and He, Qianxi and Zhou, Xunzhe and Liang, Jiaqing and Xiao, Yanghua},
  booktitle={Proceedings of the AAAI},
  volume={38},
  number={16},
  pages={18188--18196},
  year={2024}
}

@article{liu2023audioldm,
  title={Audioldm: Text-to-audio generation with latent diffusion models},
  author={Liu, Haohe and Chen, Zehua and Yuan, Yi and Mei, Xinhao and Liu, Xubo and Mandic, Danilo and Wang, Wenwu and Plumbley, Mark D},
  journal={arXiv preprint arXiv:2301.12503},
  year={2023}
}

@article{zheng2023guided,
  title={Guided flows for generative modeling and decision making},
  author={Zheng, Qinqing and Le, Matt and Shaul, Neta and Lipman, Yaron and Grover, Aditya and Chen, Ricky TQ},
  journal={arXiv preprint arXiv:2311.13443},
  year={2023}
}

@article{yang2023uniaudio,
  title={Uniaudio: An audio foundation model toward universal audio generation},
  author={Yang, Dongchao and Tian, Jinchuan and Tan, Xu and Huang, Rongjie and Liu, Songxiang and Chang, Xuankai and Shi, Jiatong and Zhao, Sheng and Bian, Jiang and Wu, Xixin and others},
  journal={arXiv preprint arXiv:2310.00704},
  year={2023}
}

@inproceedings{clifton2020100,
  title={100,000 podcasts: A spoken English document corpus},
  author={Clifton, Ann and Reddy, Sravana and Yu, Yongze and Pappu, Aasish and Rezapour, Rezvaneh and Bonab, Hamed and Eskevich, Maria and Jones, Gareth and Karlgren, Jussi and Carterette, Ben and others},
  booktitle={Proceedings of the 28th International Conference on Computational Linguistics},
  pages={5903--5917},
  year={2020}
}

@misc{parakeet-tts,
  author       = {Darefsky, Jordan and Zhu, Ge and Duan, Zhiyao},
  title        = {Parakeet},
  year         = {2024},
  month        = {May},
  day          = {12},
  howpublished = {\url{https://jordandarefsky.com/blog/2024/parakeet/}},
  note         = {Accessed: 2025-09-15}
}

@article{emo2vec,
  title={emotion2vec: Self-Supervised Pre-Training for Speech Emotion Representation},
  author={Ma, Ziyang and Zheng, Zhisheng and Ye, Jiaxin and Li, Jinchao and Gao, Zhifu and Zhang, Shiliang and Chen, Xie},
  journal={arXiv preprint arXiv:2312.15185},
  year={2023}
}

@article{utmos,
  title={UTMOS: UTokyo-SaruLab System for VoiceMOS Challenge 2022},
  author={Saeki, Takaaki and Xin, Detai and Nakata, Wataru and Koriyama, Tomoki and Takamichi, Shinnosuke and Saruwatari, Hiroshi},
  journal={Proceedings of the Interspeech 2022},
  year={2022},
  publisher={ISCA}
}

@inproceedings{whisper,
  title={Robust speech recognition via large-scale weak supervision},
  author={Radford, Alec and Kim, Jong Wook and Xu, Tao and Brockman, Greg and McLeavey, Christine and Sutskever, Ilya},
  booktitle={Proceedings of the ICML 2023},
  pages={28492--28518},
  year={2023},
  organization={PMLR}
}

@article{cho2024emosphere,
  title={Emosphere-tts: Emotional style and intensity modeling via spherical emotion vector for controllable emotional text-to-speech},
  author={Cho, Deok-Hyeon and Oh, Hyung-Seok and Kim, Seung-Bin and Lee, Sang-Hoon and Lee, Seong-Whan},
  journal={arXiv preprint arXiv:2406.07803},
  year={2024}
}

@article{lee2025hierspeech++,
  title={Hierspeech++: Bridging the gap between semantic and acoustic representation of speech by hierarchical variational inference for zero-shot speech synthesis},
  author={Lee, Sang-Hoon and Choi, Ha-Yeong and Kim, Seung-Bin and Lee, Seong-Whan},
  journal={IEEE Transactions on Neural Networks and Learning Systems},
  year={2025},
  publisher={IEEE}
}

@article{hussain2025koel,
  title={Koel-tts: Enhancing llm based speech generation with preference alignment and classifier free guidance},
  author={Hussain, Shehzeen and Neekhara, Paarth and Yang, Xuesong and Casanova, Edresson and Ghosh, Subhankar and Desta, Mikyas T and Fejgin, Roy and Valle, Rafael and Li, Jason},
  journal={arXiv preprint arXiv:2502.05236},
  year={2025}
}

@inproceedings{jing2025enhancing,
  title={Enhancing emotional text-to-speech controllability with natural language guidance through contrastive learning and diffusion models},
  author={Jing, Xin and Zhou, Kun and Triantafyllopoulos, Andreas and Schuller, Bj{\"o}rn W},
  booktitle={Proceedings of the ICASSP 2025},
  pages={1--5},
  year={2025},
  organization={IEEE}
}

@article{ma2025review,
  title={A review of human emotion synthesis based on generative technology},
  author={Ma, Fei and Xie, Yifan and Li, Yukan and He, Ying and Zhang, Yi and Ren, Hongwei and Liu, Zhou and Yao, Wei and Ren, Fuji and Yu, Fei Richard and others},
  journal={IEEE Transactions on Affective Computing},
  year={2025},
  publisher={IEEE}
}

@article{Qwen3-TTS,
  title={Qwen3-TTS Technical Report},
  author={Hangrui Hu and Xinfa Zhu and Ting He and Dake Guo and Bin Zhang and Xiong Wang and Zhifang Guo and Ziyue Jiang and Hongkun Hao and Zishan Guo and Xinyu Zhang and Pei Zhang and Baosong Yang and Jin Xu and Jingren Zhou and Junyang Lin},
  journal={arXiv preprint arXiv:2601.15621},
  year={2026}
}

@inproceedings{dnsmos,
  author={Chandan K. A. Reddy and Vishak Gopal and Ross Cutler},
  title={{DNSMOS: A Non-Intrusive Perceptual Objective Speech Quality metric to evaluate Noise Suppressors}},
  year={2021},
  booktitle={Proc. ICASSP 2021},
  pages={6493--6497},
  doi={10.1109/ICASSP39728.2021.9413981}
}

@inproceedings{nisqa,
  title     = {{NISQA: A Deep CNN-Self-Attention Model for Multidimensional Speech Quality Prediction with Crowdsourced Datasets}},
  author    = {Gabriel Mittag and Babak Naderi and Assmaa Chehadi and Sebastian Möller},
  year      = {2021},
  booktitle = {{Interspeech 2021}},
  pages     = {2127--2131},
  doi       = {10.21437/Interspeech.2021-299},
  issn      = {2958-1796},
}

@inproceedings{DPO,
  title={Direct preference optimization: Your language model is secretly a reward model},
  author={Rafailov, Rafael and Sharma, Archit and Mitchell, Eric and Manning, Christopher D and Ermon, Stefano and Finn, Chelsea},
  booktitle={Advances in Neural Information Processing Systems},
  volume={36},
  pages={53728--53741},
  year={2023}
}

@inproceedings{emodpo,
  title={Emo-dpo: Controllable emotional speech synthesis through direct preference optimization},
  author={Gao, Xiaoxue and Zhang, Chen and Chen, Yiming and Zhang, Huayun and Chen, Nancy F},
  booktitle={ICASSP 2025-2025 IEEE International Conference on Acoustics, Speech and Signal Processing (ICASSP)},
  pages={1--5},
  year={2025},
  organization={IEEE}
}

@inproceedings{dpo-tts-llm,
  title={Preference alignment improves language model-based tts},
  author={Tian, Jinchuan and Zhang, Chunlei and Shi, Jiatong and Zhang, Hao and Yu, Jianwei and Watanabe, Shinji and Yu, Dong},
  booktitle={ICASSP 2025-2025 IEEE International Conference on Acoustics, Speech and Signal Processing (ICASSP)},
  pages={1--5},
  year={2025},
  organization={IEEE}
}

@misc{betaCFG,
      title={Classifier-free Guidance with Adaptive Scaling}, 
      author={Dawid Malarz and Artur Kasymov and Maciej Zięba and Jacek Tabor and Przemysław Spurek},
      year={2025},
      eprint={2502.10574},
      archivePrefix={arXiv},
      primaryClass={cs.CV},
      url={https://arxiv.org/abs/2502.10574}, 
}

@article{dynamicCFG,
  title={Dynamic classifier-free diffusion guidance via online feedback},
  author={Papalampidi, Pinelopi and Wiles, Olivia and Ktena, Ira and Shtedritski, Aleksandar and Bugliarello, Emanuele and Kajic, Ivana and Albuquerque, Isabela and Nematzadeh, Aida},
  journal={arXiv preprint arXiv:2509.16131},
  year={2025}
}

\end{document}